\documentclass{mva_style}
\usepackage{graphicx}

\finalcopy 

\newcommand{\CO}[1]{}

\begin{document}

\newcommand{\FIG}[3]{
\begin{minipage}[b]{#1cm}
\begin{center}
\includegraphics[width=#1cm]{#2}\vspace*{-2mm}\\
{\scriptsize #3}
\end{center}
\end{minipage}
}

\newcommand{\FIGR}[3]{
\begin{minipage}[b]{#1cm}
\begin{center}
\includegraphics[width=#1cm, angle=-90]{#2}\vspace*{-2mm}\\
{\scriptsize #3}
\end{center}
\end{minipage}
}

\newcommand{\FIGpng}[5]{
\begin{minipage}[b]{#1cm}
\begin{center}
\includegraphics[bb=0 0 #4 #5, clip, width=#1cm]{#2}\vspace*{-1mm}
{\scriptsize #3}
\vspace*{1mm}
\end{center}
\end{minipage}
}

\newcommand{\FIGRpng}[5]{
\begin{minipage}[b]{#1cm}
\begin{center}
\includegraphics[bb=0 0 #4 #5, angle=-90,clip,width=#1cm]{#2}\vspace*{1mm}
{\scriptsize #3}
\vspace*{1mm}
\end{center}
\end{minipage}
}

\newcommand{\tabA}{
\begin{table*}[t]
  \caption{Experimental result.}
  \begin{center}
    \begin{tabular}{l | r r r r}
      \hline
      \hline
      {Strategy} & {\#Class} & 
      {top-5 SR} & {top-1 NSR} & {Loss} \\
      \hline
      time & 728   & 31.6\%   & & 3.97 \\
      location & 728   & 33.6\%   & & 3.92 \\
      time-apperance & 675 & 41.0\%  & 0.938\% & 3.71 \\
      location-appearance & 743   & 39.4\% & 1.664\% & 3.65 \\
      \hline
      \hline
    \end{tabular}
    \label{tab:A}
  \end{center}
\end{table*}
}

\newcommand{\figA}{
\begin{figure}[t]
  \begin{center}
    \includegraphics[width=80mm, bb=0 0 546 316]{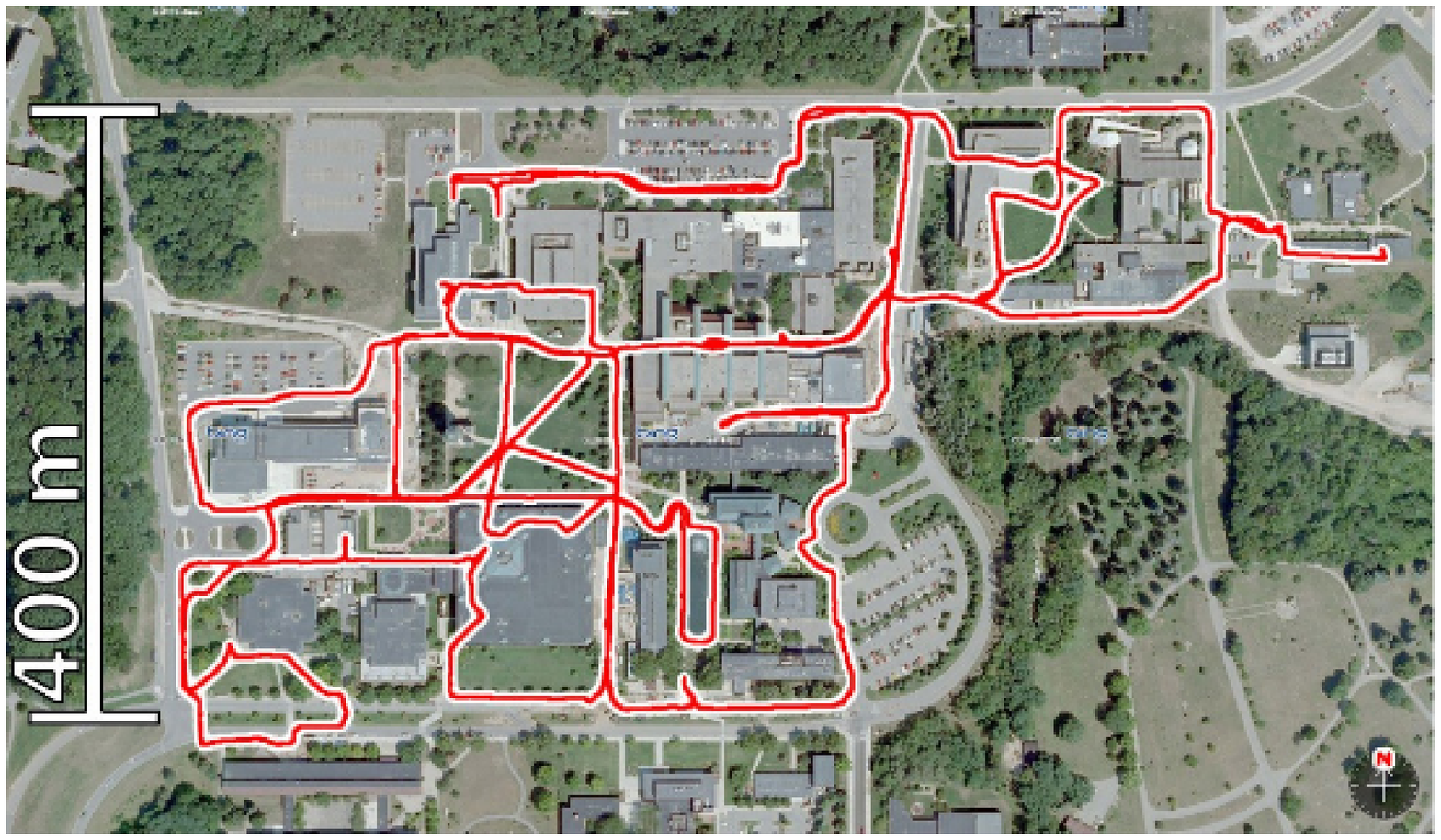}
  \end{center}
  \caption{The experimental environment and an example robot trajectory.}
  \label{fig:A}
\end{figure}
}

\newcommand{\figB}{
\begin{figure*}[t]
	\begin{center}
 \includegraphics[width=170mm, bb=0 0 1789 1049]{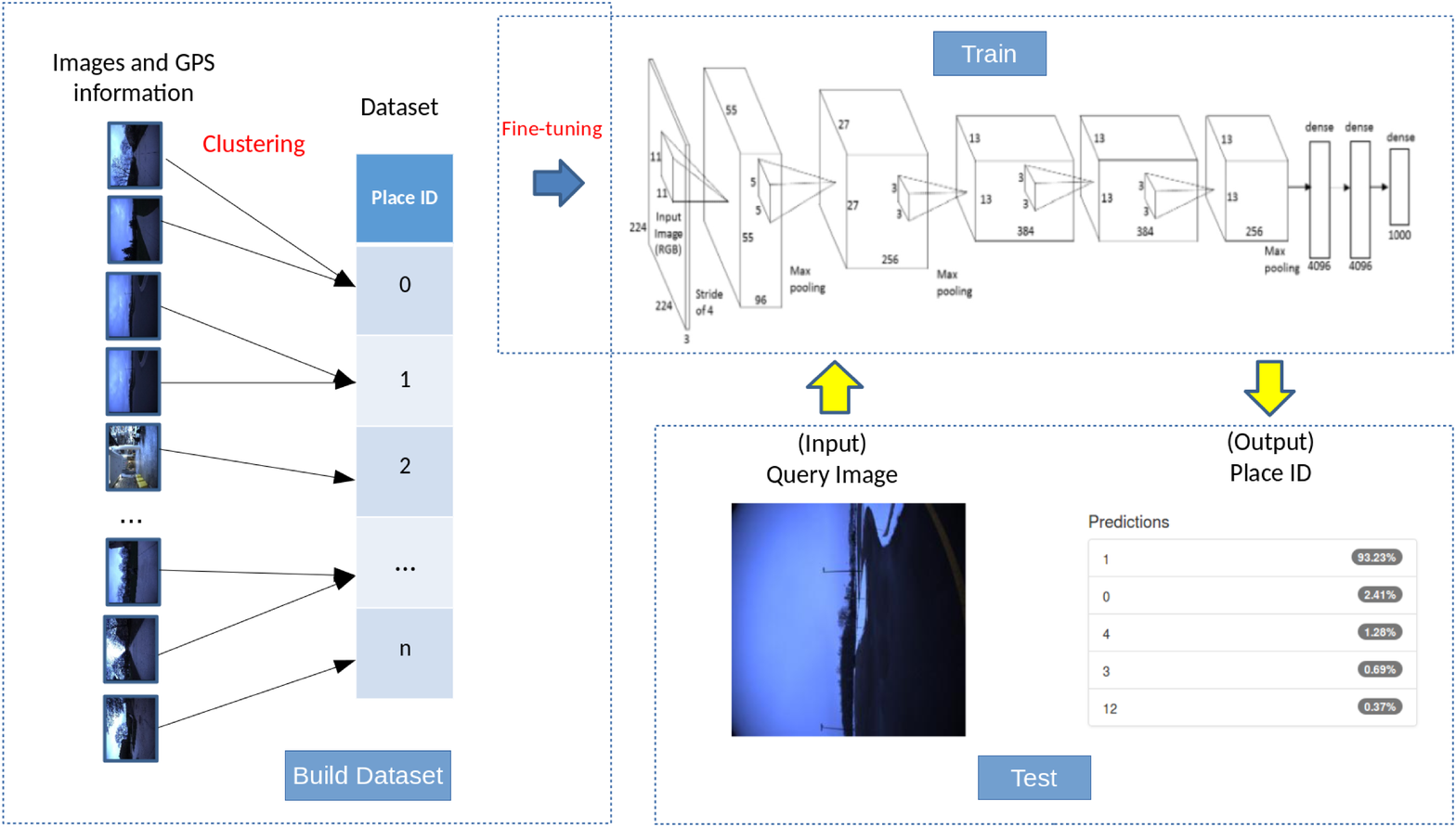}
	\end{center}
	\caption{System overview.}
	\label{fig:B}
\end{figure*}
}

\newcommand{\figC}{
\begin{figure*}[t]
  \begin{center}
\hspace*{-6mm}%
\FIG{8}{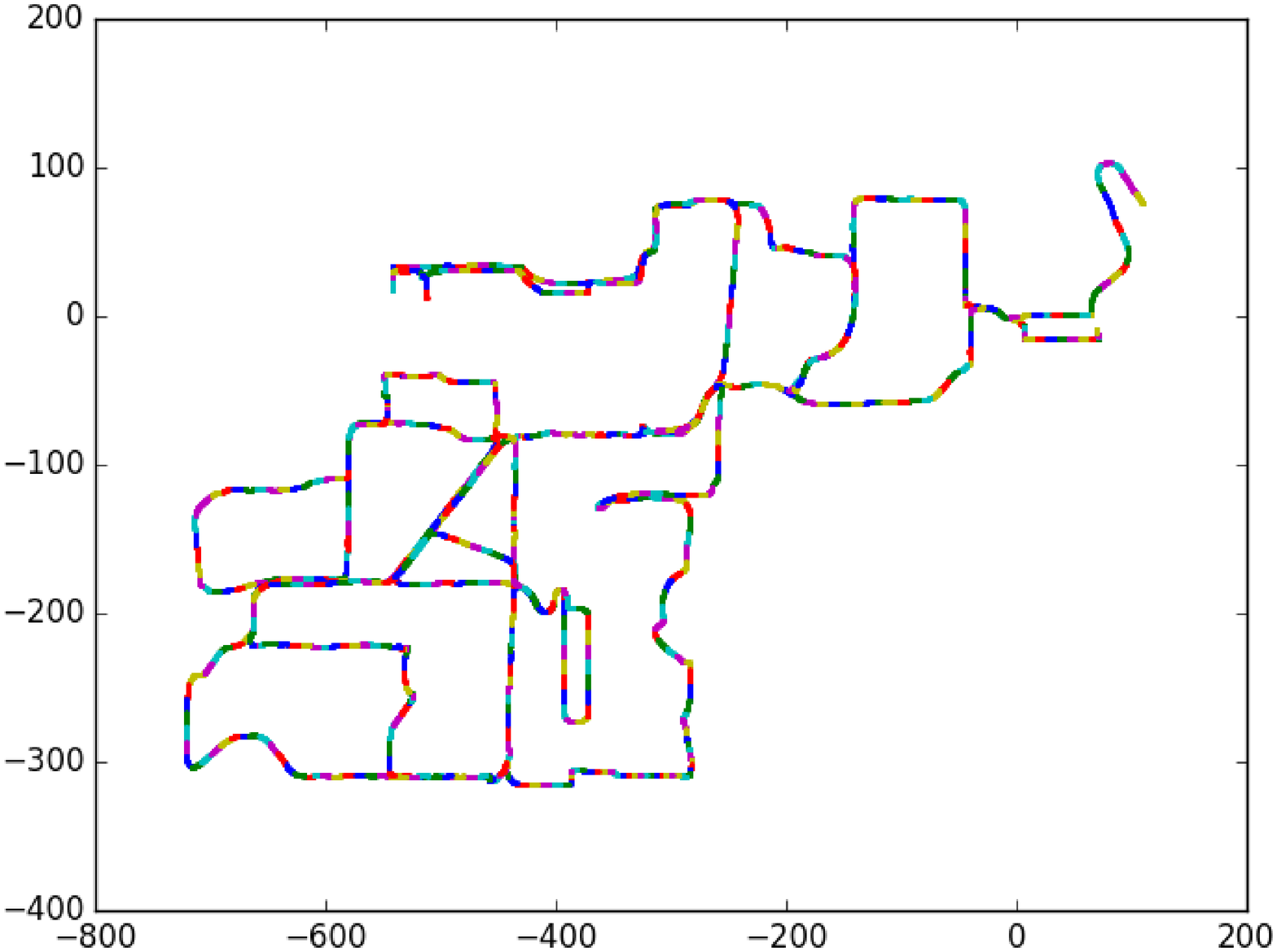}{time cue}%
\FIG{8}{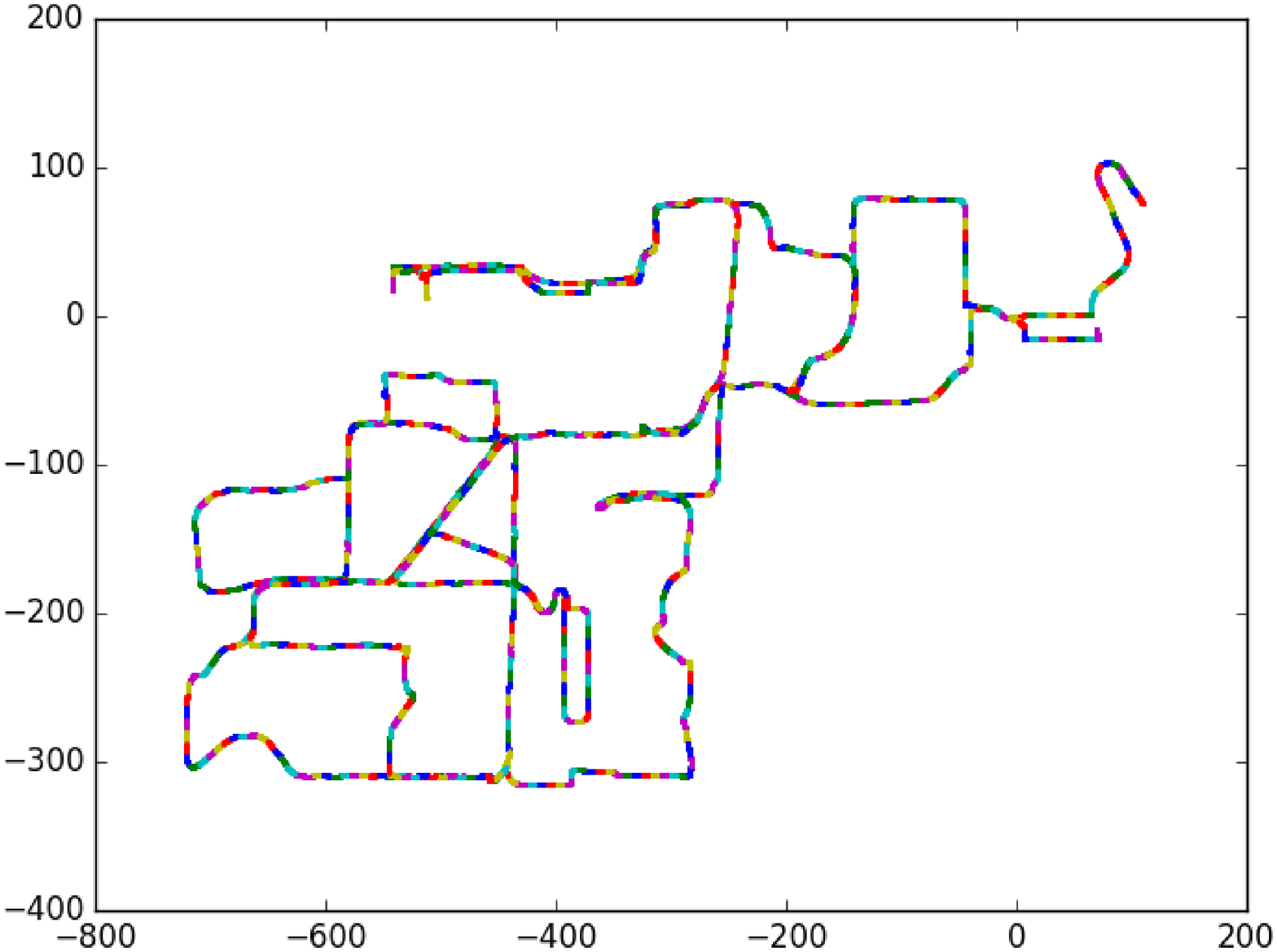}{location cue}\\
\FIG{8}{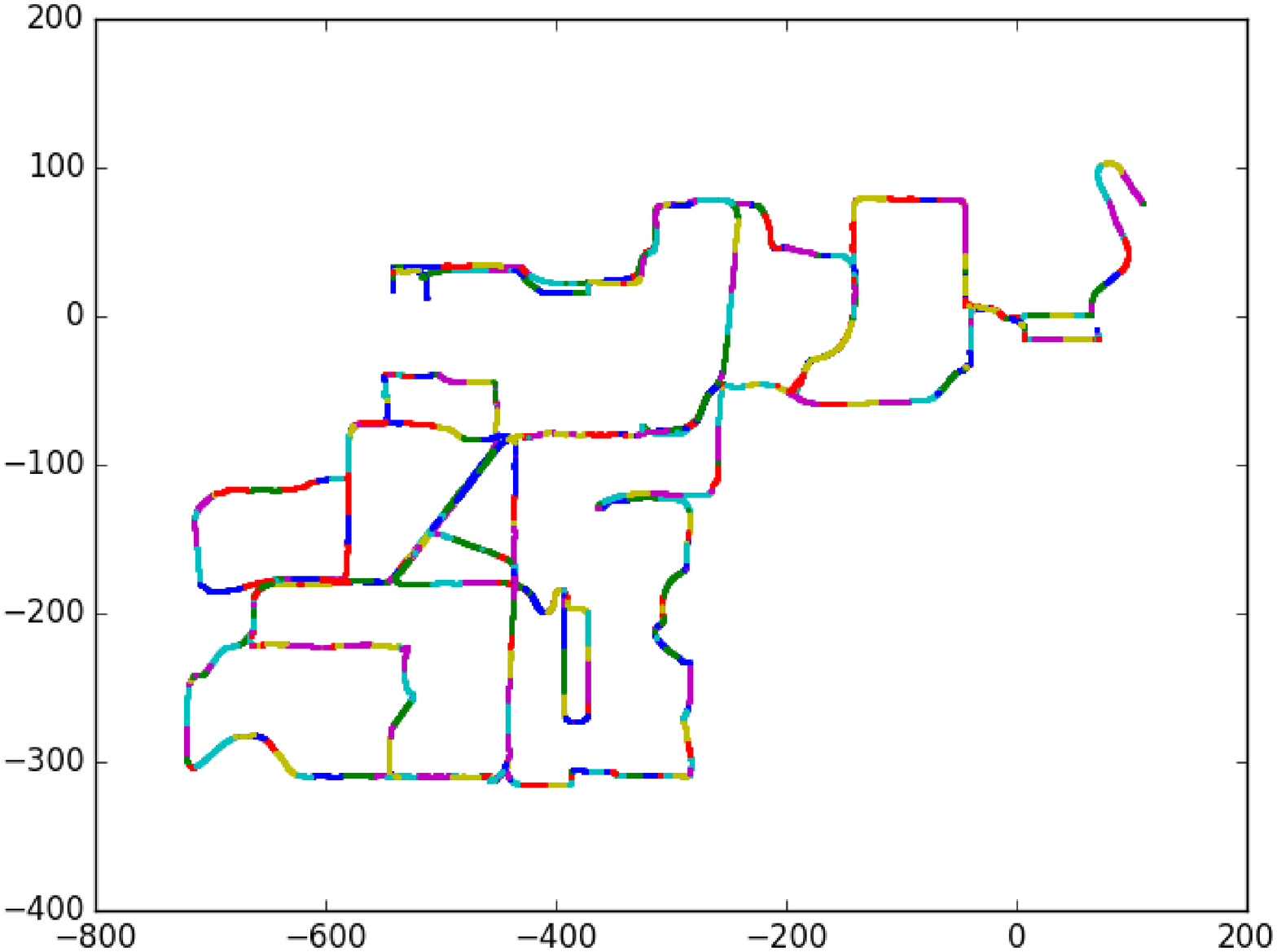}{time-appearance cue}%
\FIG{8}{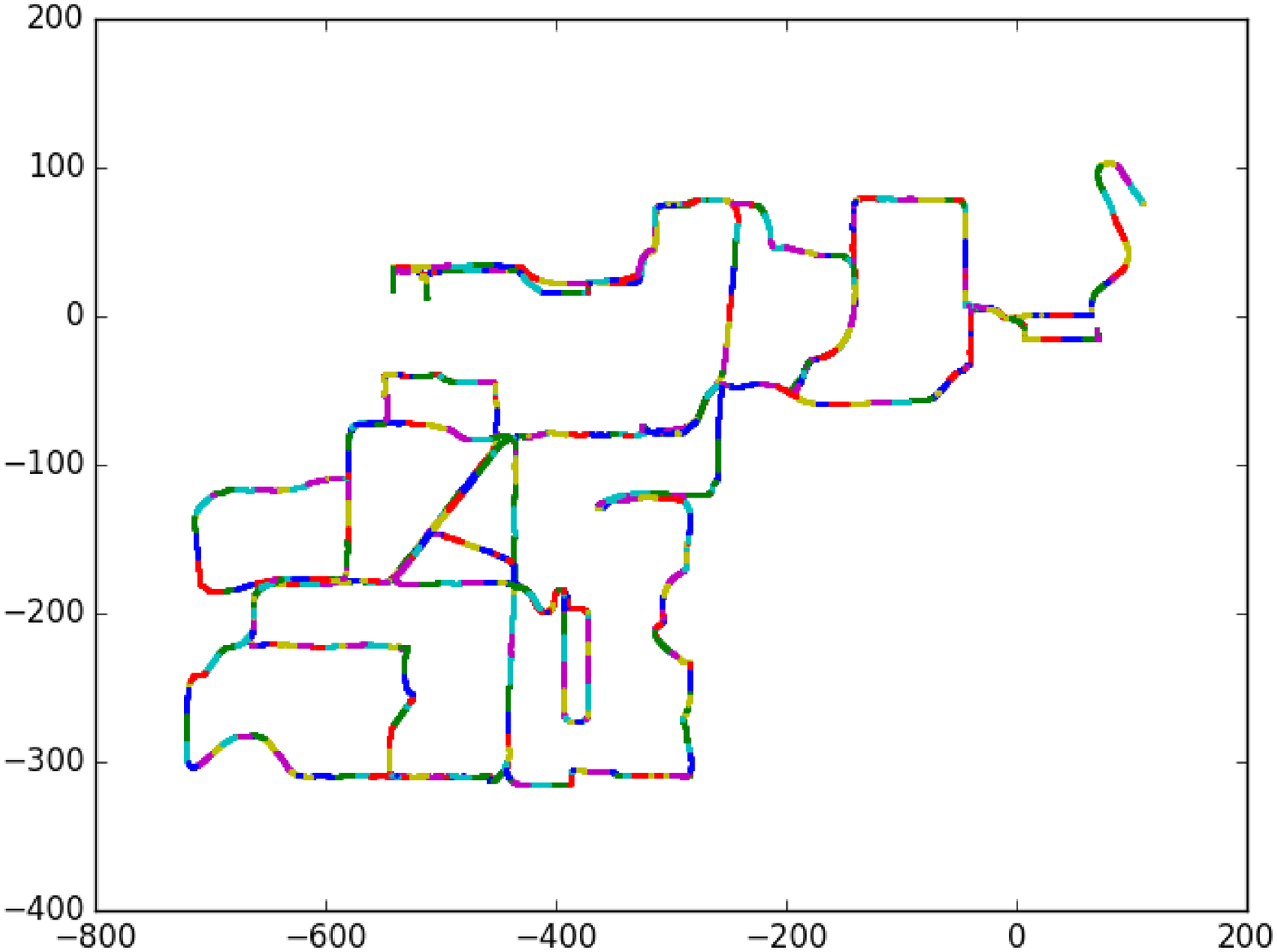}{location-appearance cue}
\caption{Example results for workspace partitioning.}
\label{fig:C}
\end{center}
\end{figure*}
}

\newcommand{\figD}{
\begin{figure}[t]
	\begin{center}
\FIGR{8}{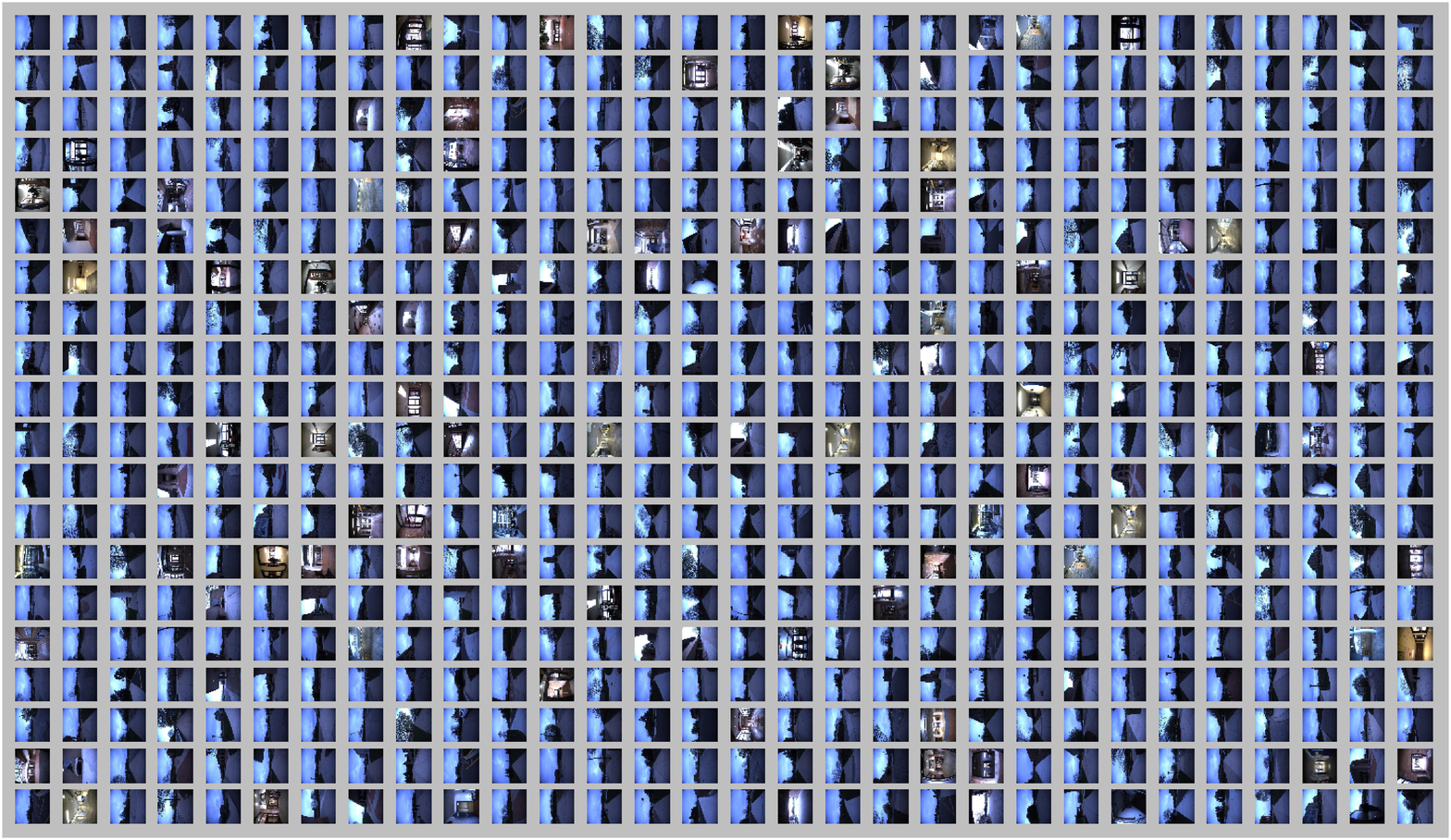}{}
	\end{center}
	\caption{Visual image for each class.}
	\label{fig:D}
\end{figure}
}

\title{
Unsupervised Place Discovery for Visual Place Classification
}
\author{
  ~\\ Fei Xiaoxiao ~~~~~~~ Tanaka Kanji ~~~~~~~ Inamoto Kouya\\
  ~\\ Univ. of FUKUI\\
  ~\\ 3-9-1, bunkyo, fukui, fukui, Japan\\
  ~\\ {\tt e-mail tnkknj@u-fukui.ac.jp}\\
}

\maketitle

\section*{\centering Abstract}
\textit{
In this study, we explore the use of deep convolutional neural networks (DCNNs) in visual place classification for robotic mapping and localization. An open question is how to partition the robot's workspace into places to maximize the performance (e.g., accuracy, precision, recall) of potential DCNN classifiers. This is a chicken and egg problem: If we had a well-trained DCNN classifier, it is rather easy to partition the robot's workspace into places, but the training of a DCNN classifier requires a set of pre-defined place classes. In this study, we address this problem and present several strategies for unsupervised discovery of place classes (``time cue," ``location cue," ``time-appearance cue," and ``location-appearance cue"). We also evaluate the efficacy of the proposed methods using the publicly available University of Michigan North Campus Long-Term (NCLT) Dataset.
}

\section{Introduction}

Visual place classification (VPC) is a fundamental task in robotic mapping and localization \cite{1}.
In VPC, a mapper robot collects a set of training images with ground-truth viewpoint information, assigns a class label (place ID) to each image, and learns an environment map from the labeled training data.
Then, a map user robot takes a visual image without viewpoint information, and classifies it into one of the learned place classes.
In this paper, we are motivated by the recent success of deep convolutional neural network (DCNN) \cite{2} in various classification tasks, and to explore the use of a DCNN classifier as an environment map.

An open question is how to partition the robot's workspace into places. This is an important problem as the definition of place classes strongly influences performance (e.g., accuracy, precision, recall) of a VPC task. Intuitively, each place class should be defined as a continuous region in the robot's workspace with similar DCNN features. The main difficulty is a chicken and egg problem: If we had a well-trained DCNN classifier, it is rather easy to partition the robot's workspace into place regions, but the training of a DCNN classifier requires a set of pre-defined place classes. 

In this study, we formulate this problem and present several strategies. It is assumed that we are given a collection of visual images with ground-truth viewpoint as a guide, which can be independent from training and test data. The goal is to search for an effective partition of the workspace into places to maximize performance of potential DCNN classifiers. We propose to use three different types of information: time cue, location cue, and appearance cue that is available from the pre-trained DCNN. We then present four different strategies for workspace partitioning by combining them: ``time cue," ``location cue," ``time-appearance cue," and ``location-appearance cue". Finally, we evaluate the efficacy of the proposed methods using the publicly available University of Michigan North Campus Long-Term (NCLT) Dataset \cite{3}.

This work is inspired by our previous work on the use of DCNNs in visual place classification \cite{kanji2016self}. In \cite{kanji2015cross}, we address the issue of unsupervised place discovery. In it, we view places (i.e., mapped images) as independent classes, and for each class, we form a class-specific set of training features, by mining the visual experience to find the relevant library features that effectively explain the input scenes. In contrast, this study focuses on the use of DCNNs for unsupervised place discovery. The issue has not been explored in the above works.

\figB

\section{Approach}

\subsection{System Overview}

Fig. \ref{fig:B} shows an overview of our approach. We assume a typical supervised classification framework for VPC. The entire VPC framework consists of two phases: (1) training, and (2) testing. The training phase (``Train" in Fig. \ref{fig:B}) takes as input a set of labeled training images for each place class and trains a classifier that classifies an image into one of the pre-defined place classes. We represent the place class ID with a label. The testing phase (``Test" in Fig. \ref{fig:B}) takes as input a novel unseen image (``Query image" in Fig. \ref{fig:B}) and predicts its place class by using the trained classifier. Our use of DCNN for learning and testing follows typical transfer learning \cite{17}, where the DCNN is pre-trained on Big Data and then fine-tuned to adapt to the target domain \cite{2}. Our experimental system is based on Alexnet pre-trained on the ImageNet LSVRC-2012 dataset and fine-tuned (``Fine-tuning" in Fig. \ref{fig:B}) on the relatively small training dataset (``Dataset" in Fig. \ref{fig:B}), which our algorithm creates from the NCLT Dataset (``Build dataset" in Fig. \ref{fig:B}). 

Although our approach is sufficiently general and applicable to various types of environments (e.g., indoor and outdoor) and sensor modalities, in experiments, we focus on the NCLT Dataset \cite{3}. The NCLT Dataset is a large scale, long-term autonomy dataset for robotics research collected on the University of Michigan's North Campus by a Segway robotic platform.  The Segway was outfitted with: a Ladybug3 omnidirectional camera, a Velodyne HDL-32E 3D lidar, two Hokuyo planar lidars, an inertial measurement unit (IMU), a single-axis fiber optic gyro (FOG), a consumer grade global positioning system (GPS), and a real-time kinematic (RTK) GPS. The data we used in our research includes image and navigation data from the NCLT Dataset. The image data is from the front facing camera (camera\#5) of the Ladybug3 omnidirectional camera. Fig. \ref{fig:A} shows a bird's eye view of the experimental environment and an example robot trajectory.

Every test image is assigned a place label. For every test image, a set of training images with sufficiently similar viewing directions is selected. From these training image sets, the one with nearest location is selected and its place label is assigned to the test image. We set the threshold on orientation similarity to 20 deg. If the distance from the ground-truth viewpoint of a test image to that of any training image is larger than a pre-defined threshold of 18 m, the image is considered invalid as a test image and not used in our classification experiments.

\figA

\figC

\subsection{Problem Formulation}

The workspace partitioning problem is formulated as follows. The goal is to partition the robot's workspace into continuous place regions to maximize the performance of potential DCNN classifiers. The input is a set of images and viewpoints collected by the mobile robot in the target environment, which can be independent from training and testing data. The output is a set of clusters of images, and we assign a place ID to each cluster. It is natural that the workspace partitioning takes place prior to the training phases, and influences performance of both the training and classification.

\subsection{Performance Index}

We evaluate the performance of a workspace partitioning algorithm using simulated VPC tasks. The evaluation procedure consists of three distinct steps. First, the training dataset is partitioned into $K$ clusters using the algorithm of interest. Second, the DCNN classifier (pre-trained on the ImageNet LSVRC-2012 dataset) is fine-tuned using the corresponding $K$ image sets as training data. Third, each test image in the test set is fed to the trained DCNN classifier to obtain the classification result, and the performance is evaluated by success rate (SR) in the form:
\begin{equation}
p = \frac{1}{K} \sum_{n=1}^K [a_n\in r_n], \label{eqn:eval}
\end{equation}
$n$ is the ID of test image.
$r_n$ is the top-$X$ classification result of the $n$-th test image (e.g., $X=1$, $X=5$),
which is represented by a size $X$ set of training image IDs.
$a_n$ is the ground truth classification result,
which is represented by a single training image ID.
$[\cdot]$ is an indicator function:
\begin{equation}
[a_n\in r_n] = \cases{ 1 & (if $a_n\in r_n$) \cr
0 & (otherwise) \cr
}.
\end{equation}

A limitation of the above evaluation function (\ref{eqn:eval}) is that it ignores the size $|C(r_n)|$ of image sets that belong to each image class $C(r_n)$. From the perspective of robotic mapping and localization, the accuracy of the VPC should reflect the cluster size $|C(r_n)|$, as the robot's final goal is to localize a single test image rather than a cluster of test images. To address this issue, we also introduce a normalized version of the success rate (NSR) in the form:
\begin{equation}
p = \frac{1}{K} \sum_{n=1}^K \frac{[a_n\in r_n]}{|C(r_n)|}.
\end{equation}
Note that $|C(x_n)|$ serves as a regularizer to avoid meaningless solutions in which all the training images belong to a single place class and other place classes are empty. Intuitively, $p$ represents the probability of a given image being correctly classified into the ground-truth place class.

\subsection{Workspace Partitioning Strategies}

We developed four different strategies for workspace partitioning. Fig. \ref{fig:C} shows examples of place classes found by each strategy. In the figure, different colors indicate different place classes. We can see that all the strategies create a set of clusters with similar cluster sizes, although the performance difference between them is significant as shown in the experimental section, Section \ref{sec:exp}. 

The first is a simple time cue strategy. It partitions a sequence of images into classes by their time stamps or image IDs. Thus, the partitioning result is a set of clusters with $K-1$ intervals of approximately equal duration. This strategy is based on an observation that images with similar time stamps are expected to have visually similar appearance, since they are collected by a mapper robot that navigates through a continuous trajectory in the environment, and such clusters of images are expected to be a good training set for a DCNN classifier. Obviously, this simple strategy has many limitations. Particularly, it is not robust if the robot's moving speed varies. Also, it does not take advantage of any appearance features that are available from the pre-trained DCNN.

The second strategy is location cue strategy. This strategy is different from the time cue strategy only in that it partitions a sequence of images not by their time stamps, but by their travel distance along the trajectory. Thus, the partitioning result is a set of clusters with $K-1$ intervals of approximately equal travel distances. Note that the location cue strategy does require the information of travel distance along the trajectory, which is readily available given the ground-truth viewpoint information. This strategy is robust against variance in the robot's moving speed but still does not take advantage of appearance information from the pre-trained DCNN.

The third strategy is location-appearance cue strategy. The basic idea is to augment the location cue strategy by using the available pre-trained DCNN classifier as a guide. We use the 6-th layer from the pre-trained DCNN as the image representation because it has shown excellent performance in the image classification task in \cite{4}. The workspace partitioning procedure is as follows. (1) Images are represented by 4,096 dimensional 6-th layer features from the DCNN. (2) They are fed to k-means clustering to obtain $K$ image clusters. (3) For each cluster, we perform the location cue strategy to partition the cluster into sub-clusters.

The fourth strategy is time-appearance cue strategy. The basic idea is to augment the time cue strategy by using the available pre-trained DCNN classifier as a guide. The basic concept of the augmentation is similar to that of the location-appearance cue strategy, but different in that we perform the time cue strategy (instead of the location cue strategy) in the step3 of the procedure.

\figD

\tabA

\section{Experiments}\label{sec:exp}

We evaluated the proposed framework for workspace partitioning on real VPC tasks. Four different strategies for workspace partitioning were considered: time, location, location-appearance, and time-appearance. For training data, we used the dataset of ``March 31st 2012," in which the total travel distance is 6.0km, the time is midday, the environmental condition is cloudy, no foliage and no snow. The training data consists of images and viewpoint information that is available from the NCLT Dataset. For testing data, we used the dataset of ``Aug 4th 2012," in which the travel distance is 5.5km, the time is morning, the environmental condition is sunny, foliage and no snow. The test data consists of images, and the available viewpoint information is used for ground-truth prediction. Fig. \ref{fig:D} shows examples of images belonging to individual clusters in the case of time-appearance strategy. Shown are representative images from 600 clusters that were randomly sampled from the $K$ clusters.

The training data provided by each strategy was fed to transfer learning (i.e., fine-tuning) of the DCNN that is pre-trained on the Big Data (i.e., the ImageNet LSVRC-2012 dataset). The classification function in the pre-trained DCNN is a softmax classifier that computes the likelihood over 1,000 classes of the ImageNet dataset. To fine-tune the DCNN, we changed the number of the softmax classifiers at the top layer with the number of place classes. Then, the DCNN parameters were fine-tuned on new training datasets. After the fine-tuning, we evaluated the performance of the DCNN on the test set in terms of accuracy. In the experiment, we changed the softmax classifier with a new value that is equal to the classes of training datasets. 

Table \ref{tab:A} shows the loss and accuracy of the test datasets. We can see that the time-appearance cue strategy outperformed the time cue, location cue, and location-appearance cue strategies. We also see that the location cue strategy outperformed the time cue strategy. The reason may be that the time cue strategy does not consider that the mapper robot moves with variable and it often fails to partition the workspace into equal-size sub-regions. Overall, the time-appearance and location-appearance strategies outperformed the other two. It could be said that appearance information from the pre-trained DCNN provides an effective cue to further improve the time cue and location cue strategies. Finally, the success rate of these strategies is sufficiently high considering the fact that the number of possible places is large (e.g., 675).

\section{Conclusions \& Future Works}

In this study, we explored the use of deep convolutional neural networks (DCNNs) in visual place classification (VPC) for robotic mapping and localization. It has been shown that the proposed strategies for workspace partitioning enabled effective discovery, learning and classification of place classes. Our research showed that we can use location features and appearance features to partition the robot's workspace into places, which leads to better fine-tuning of the DCNN, and improves overall performance of VPC.

\bibliographystyle{IEEEtran}
\bibliography{draft}

\end{document}